\documentclass[lettersize,journal]{IEEEtran}
\usepackage{amsmath,amsfonts}
\usepackage{algorithmic}
\usepackage{algorithm}
\usepackage{array}
\usepackage[caption=false,font=normalsize,labelfont=sf,textfont=sf]{subfig}
\usepackage{textcomp}
\usepackage{stfloats}
\usepackage{url}
\usepackage{verbatim}
\usepackage{graphicx}
\usepackage{cite}
\usepackage{graphicx}
\usepackage{amsmath}
\usepackage{amssymb}
\usepackage{booktabs}
\usepackage{mathtools}
\usepackage{hhline}
\usepackage{subfig}
\usepackage{multirow}
\usepackage[pagebackref,breaklinks,colorlinks]{hyperref}
\usepackage{caption}
\begin{document}

\title{Interaction Transformer for Human Reaction Generation}

\author{Baptiste~Chopin,
Hao~Tang,
Naima~Otberdout,
Mohamed~Daoudi,~\IEEEmembership{Senior Member,~IEEE,}
Nicu~Sebe,~\IEEEmembership{Senior Member,~IEEE}
\IEEEcompsocitemizethanks{ \IEEEcompsocthanksitem B. Chopin is with Univ. Lille, CNRS, Centrale Lille, UMR 9189 CRIStAL, F-59000 Lille, France. E-mail:. baptiste.chopin@univ-lille.fr
\IEEEcompsocthanksitem Hao Tang is with the Department of Information Technology and Electrical Engineering, ETH Zurich, Zurich 8092, Switzerland. E-mail: hao.tang@vision.ee.ethz.ch
\IEEEcompsocthanksitem N. Otberdout is with Ai movement - University Mohammed VI Polytechnic, Rabat, Morocco, E-mail: naima.otberdout@um6p.ma
\IEEEcompsocthanksitem M. Daoudi is with IMT Nord Europe, Institut Mines-Télécom, Univ. Lille, Centre for Digital Systems, F-59000 Lille, France, and Univ. Lille, CNRS, Centrale Lille, Institut Mines-Télécom, UMR 9189 CRIStAL, F-59000 Lille, France, E-mail: mohamed.daoudi@imt-nord-europe.fr
\IEEEcompsocthanksitem Nicu Sebe is with the Department of Information Engineering and Computer Science, University of Trento, Trento 38123, Italy.
E-mail: niculae.sebe@unitn.it }
\thanks{Manuscript received April 19, 2021; revised August 16, 2021.}}



\maketitle

\begin{abstract}
We address the challenging task of human reaction generation, which aims to generate a corresponding reaction based on an input action. Most of the existing works do not focus on generating and predicting the reaction and cannot generate the motion when only the action is given as input. To address this limitation, we propose a novel interaction Transformer (InterFormer) consisting of a Transformer network with both temporal and spatial attention. Specifically, temporal attention captures the temporal dependencies of the motion of both characters and of their interaction, while spatial attention learns the dependencies between the different body parts of each character and those which are part of the interaction. Moreover, we propose using graphs to increase the performance of spatial attention via an interaction distance module that helps focus on nearby joints from both characters. Extensive experiments on the SBU interaction, K3HI, and DuetDance datasets demonstrate the effectiveness of InterFormer. Our method is general and can be used to generate more complex and long-term interactions. We also provide videos of generated reactions and the code with pre-trained models at \href{https://github.com/CRISTAL-3DSAM/InterFormer}{\textcolor{magenta}{github.com/CRISTAL-3DSAM/InterFormer}}.
\end{abstract}


\section{Introduction}

\begin{figure}[!t]
    \centering
    \includegraphics[width=1\linewidth]{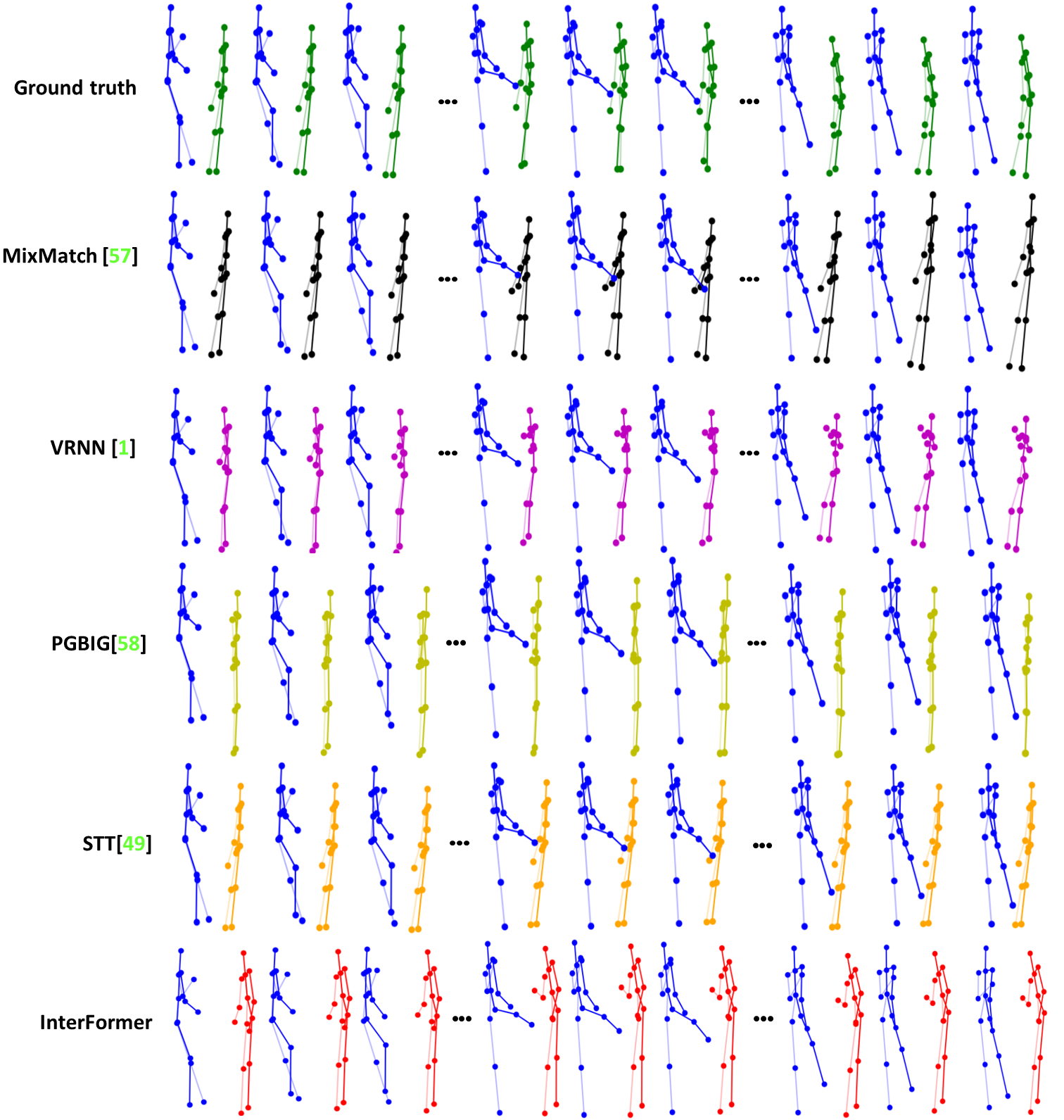}
    \caption{\textbf{Example of reaction generation.} In blue the action motion is used as a condition. In other colors, the reaction is either from the ground truth or generated by the different models. Example from the kicking class of the SBU dataset. Our model generates a more realistic motion than the competing approaches.}
    \label{fig:intro_SBU_kick}
\end{figure}

Modeling the dynamics of human motion is at the core of many applications in computer vision and robotics. Most works on human motion generation ignore human interactions and focus rather on the generation of actions of a single person. In addition, only a few works investigating human interaction generation~\cite{multimodal_interaction} look at the reaction generation problem. What makes human reaction generation a challenging problem is the non-linearity in the temporal evolution of human motion and the two sources that condition the motion: the action and its corresponding reaction. The first issue arises because human motion is generally performed at varying evolution rates. In other words, a person performing the same activity will go roughly through the same stages but at slightly different rates every time. In addition, as stated by~\cite{duetdance}, unlike simple actions such as walking or running, complex human interactions such as duet dancing generate highly complex pose sequences operating close to the limit of human kinematics with very low periodicity. The second issue arises because the same action can have a different reaction depending on the interaction context, e.g., when reacting to a punch depending on the position, one can react more or less strongly. These two issues make the problem of reaction generation and evaluation challenging. Several questions arise as we try to tackle this challenge. How to translate action to reaction? How to model the long-term sequence? How to represent a complex action-reaction sequence? 



\begin{figure*}[!t]
    \centering
    \includegraphics[width=1\linewidth]{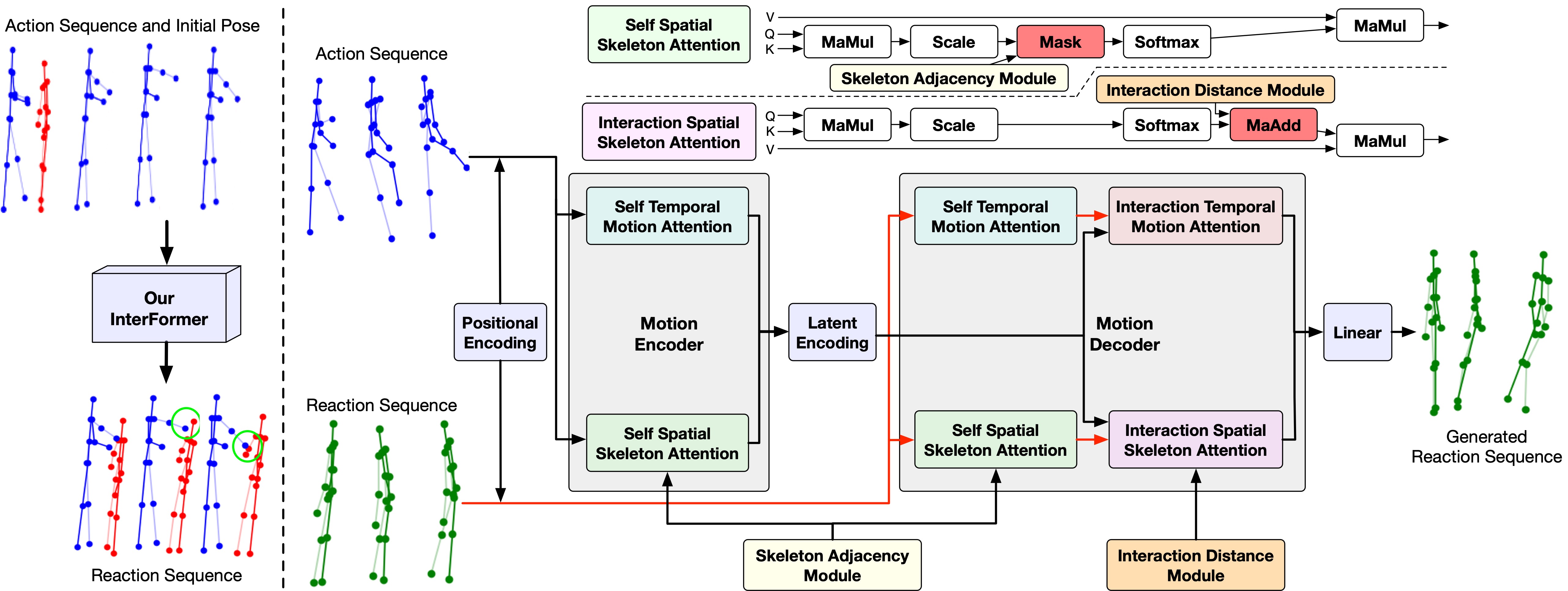}
    \caption{\textbf{Left:} InterFormer during testing: given an action sequence (blue) and the first frame of a reaction sequence (red), we generate the full reaction sequence. We predict one frame at a time based on the previously generated frames. \textbf{Right:} Overview of InterFormer during training. The motion encoder takes an action sequence as the input and outputs a latent encoding. The motion decoder takes as inputs the reaction sequence corresponding to the action sequence and the latent encoding from the encoder. The motion decoder outputs a generated reaction sequence. Both the encoder and decoder contain several attention modules. 
    \textbf{Top Right:} The skeleton adjacency and interaction distance modules interact directly with spatial attention.}
    \label{fig:overview}
\end{figure*}

Our goal is to learn the reaction from a training sequence of actions and reactions by using Transformer architectures. The breakthroughs from Transformer networks in Natural Language Processing (NLP) domain have sparked great interest in computer vision. Transformer architectures are based on a self-attention mechanism that learns the relationships between elements of a sequence. Unlike recurrent networks that process the elements of the sequence recursively, Transformers can attend to complete sequences and thereby are able to learn spatial and temporal relationships making them a good candidate for modeling human motion. In this paper, we propose InterFormer, which with its spatial and temporal attention modules, is able not only to model the spatial and temporal dependencies in the action and in the reaction but also in the interaction between the two humans providing a solution to the two previously mentioned issues. Figure~\ref{fig:overview} (Left) shows how our InterFormer can generate a proper reaction sequence (red skeleton) by taking as input an action sequence (blue skeletons) and the initial position of the reaction sequence. Green circles highlight the reaction parts of the motion: the head goes backward in reaction to the punch; the hand is raised as the body continues to move backward to keep its balance. Figure~\ref{fig:intro_SBU_kick} shows a generated reaction from the ``kicking'' class of the SBU dataset. Our method is able to generate a proper motion.

Our major contributions are as follows:
\IEEEpubidadjcol
\begin{itemize}
\item We propose a novel Interaction Transformer framework for the challenging human reaction generation task. To the best of our knowledge, this is the first work that challenges the task of human reaction prediction given the action of the interacting skeleton using a Transformer based architecture.
\item We formulate the reaction generation problem as a translation problem, where we translate a given action of a skeleton to its corresponding reaction such that the entire interaction looks coherent and natural.
\item We adopt a graph representation for self-attention to better exploit the skeleton structure while we ignore this representation for computing the attention between the two interacting skeletons. In this case, instead of a graph representation, we exploit the distance between the interacting joints assuming that closer joints involve stronger interaction. By introducing this distance, we provide the prior knowledge that helps to model the interaction.
\item While the previous methods for interaction generation address limited and simple short-term interactions, we evaluate our method on the DuetDance dataset that provides more complex and long-term interactions.
\end{itemize}
\IEEEpubidadjcol

\section{Related Work}

\noindent \textbf{Human Action Generation.} Human action recognition and prediction from 3D skeletons is a popular topic~\cite{Kacem-Pami-2020,martinez_human_2017,Cui_2020_CVPR,devanne20143,RNMhumandyna, ZhaoACMMM2023,ChopinTBIOM2022}. Inspired by the recent advances in generative models, several works \cite{motiongen,zhenyi2020,autoconditioned,Li2021AICM} proposed human action generation models in order to generate a consecutive sequence of human motions. Recently there has been an increase in motion generation based on different modalities, \cite{yin2021graphbased} use control signals such as the global trajectory of the person to generate human motion in long-term horizons while \cite{ahuja-etal-2020-gestures} and \cite{3dconvgesture_2021} generate motion based on speech audio. Meanwhile, others use only knowledge of the past motion which allows them to work in real-time but on shorter motion \cite{martinez_human_2017,Cui_2020_CVPR,Sofianos_2021_ICCV}. However, these works only focus on the generation of individual actions. More recently, interaction prediction and generation have also been addressed \cite{multimodal_interaction,duetdance}. 
For instance, \cite{multimodal_interaction} use a multimodal variational recurrent
neural network to predict the future motion of both participants in an interaction based on pasts sequences of motion. To complement the existing dataset with interactions ~\cite{NTU,kiwon_hau3d12}, different types of complex interaction datasets have also emerged like~\cite{talkinghands} and their collection of conversational hand motions or triadic interactions~\cite{triyadic}. Some works also look at human reaction with other modalities such as walking trajectories~\cite{socialtraj} or conversational data~\cite{reactnotreact,socialsynthesis,multimodalgesture}. More recently, a lot of focus has been devoted to human pose and motion generation from text or action labels, as well as its reciprocal task ~\cite{GuoECCV22, Lucaseccv2022}. However, our approach proposes to generate and predict human motion reactions from an action. In addition, these papers focus only on one person, while our approach is dedicated to the generation of reactions in two-person interaction.
However, there are very few works on human reaction generation. In this paper, we focus on this and propose InterFormer, a novel Transformer architecture. This idea has not been investigated by any other existing work.

\noindent \textbf{Graph Representation} has been widely used for 3D classification and segmentation~\cite{qi20173d,qi2017pointnet}, visual question answering~\cite{norcliffe2018learning}, human interaction recognition~\cite{ZHU2021107920,PLIZZARI2021103219}. For instance, \cite{ZHU2021107920} proposed a Dyadic Relational Graph Convolutional Network (DR-GCN) for skeleton-based interaction recognition.  When dealing with 3D skeletons, it is natural to use a graph representation as the graph of the skeleton exists physically in the form of segments linking joints. While most works use the graph representation of the skeleton directly as an input, doing so when dealing with interaction leads to losing information. We propose to use the graph as part of the attention module to take advantage of the graph representation without losing the information about the interaction. Experiments show the effectiveness of the proposed InterFormer over existing methods.\\
\noindent \textbf{Vision Transformers.} Transformers were introduced in~\cite{transformer} as a new attention-based building block for machine translation. Because the architecture was powerful and flexible, it was quickly adapted to other natural language processing tasks like language modeling~\cite{AlRfou2019CharacterLevelLM,DevlinCLT19}. They also have recently demonstrated good performance on a broad range of tasks such as image classification \cite{kong2022spvit}, image generation~\cite{pmlr-v80-parmar18a,Esser_2021_CVPR}, object detection~\cite{Carion-eccv-2021,kong2022peeling}, human pose estimation \cite{li2022mhformer}, depth estimation~\cite{Ranftl2021,yang2021transformer}, 3D pose transfer \cite{chen2022geometry,chen2021aniformer}, and action recognition~\cite{PLIZZARI2021103219, Martinez_ICCV_2021}. Closer to our problem,  works have used Transformer to generate human motion:  \cite{petrovich21actor,2022arXiv220307706X} generate human motions based on the class labels while \cite{aksan2021spatio} use them to predict future motion based on a historical sequence. Different from these methods, we use a Transformer architecture with temporal and spatial attention for solving the reaction generation task. Generating a reaction responding to an action can be seen as a translation problem: translating from a language ``action'' to a language ``reaction''. The performance of the Transformer on natural language translation tasks and its use of temporal information is a good fit for our task of reaction generation. By adding spatial attention and graph information, we can produce a realistic reaction to an action. To the best of our knowledge, InterFormer is the first Transformer architecture used to solve the problem of human reaction generation.


\section{The Proposed Interaction Transformer}
Let us consider $P_t$ the positions of $k$ distinct joints at time $t$. Consequently, an action sequence $P$ of $T$ frames, can be described as a sequence $P{=}\{P_1, P_2, \dots, P_T\}$, where $P_t  {\in} \mathbb{R}^d$ and $d{=}3 {\times} k$,
where $P_t{=}[J_{1}(t), \dots, J_{k}(t)],$ with $k$ the number of joints in the skeleton, and  $J_{i}(t){=}[x_{i}(t), y_{i}(t), z_{i}(t)]$ the 3D coordinates of joint $i$. The goal is to generate a reaction $Y{=}\{Y_1, Y_2, \dots, Y_T\}$ a sequence of skeleton poses from $X{=}\{X_1, X_2, \dots, X_T\}$ a sequence representing the action motion. 

Our overall architecture of InterFormer is illustrated in Figure~\ref{fig:overview} and consists of four modules: a motion encoder, a motion decoder, a skeleton adjacency module, and an interaction distance module.
The motion encoder encodes the motion of the skeleton using a self-spatial skeleton and self-temporal motion attention. Both aim to find the important spatial and temporal relations within the input action motion to transmit them to the decoder. The motion decoder generates the reaction motion using the encoding from the motion encoder and consists of self-spatial skeleton attention, self-temporal motion attention, interaction spatial skeleton attention, and interaction temporal motion attention. 
Moreover, the skeleton adjacency and interaction distance modules help the different spatial attentions to focus on the most important parts of the skeletons and of the interaction.


\subsection{Motion Encoder}
The motion encoder takes as input an action sequence $X$ to which we add positional encoding defined by \cite{transformer}. This positional encoding encodes temporal information of each frame in the sequence.
Inspired by \cite{transformer} we use temporal attention to capture the temporal relationships within the motion of the skeleton. However, the motion contains both temporal and spatial information. Thus, we add a spatial attention module to complement the temporal attention to help find the spatial dependencies within the skeleton. 

\noindent \textbf{Self Spatial Skeleton Attention.}
For our self-spatial skeleton attention module, we consider each frame independently and look at the relation between the position of each joint. We use the scaled dot-product attention from~\cite{transformer}:
\begin{equation}
\label{eq:temporal attention}
\operatorname{Attention} (\mathbf{Q},\mathbf{K},\mathbf{V})=\operatorname{softmax}\left(\dfrac{\mathbf{Q}\mathbf{K}^T}{\sqrt{dim}}\right)\mathbf{V},
\end{equation}
where $\mathbf{Q}$, $\mathbf{K}$, and $\mathbf{V}$ are the query, key, and value matrices of sizes $dim {\times} |P_t|$ which contain a set of queries, keys, and values (one for each joint in the skeleton for a given frame) of sizes $dim$ which is for spatial attention $|J_1(t)|$. These queries $q_i$, keys $k_i$, and values $v_i$ are obtained by multiplying an input $a_i$, $b_i$, and $c_i$ by weight matrices $\mathbf{W}_q$, $\mathbf{W}_k$, and $\mathbf{W}_v$ of size  $dim {\times} dim$:
\begin{align}
\label{eq:weights}
    q_i&=a_i \mathbf{W}_{q},  &
    k_i&=b_i \mathbf{W}_{k},  &
    v_i&=c_i \mathbf{W}_{v}.
\end{align}

For self-attention $a_i{=}b_i{=}c_i$ and for spatial attention they represent the 3D coordinates of joint $i$ at a given time, either directly or through the value corresponding to the coordinates after going through the previous attention layers. We use the multi-head version of the attention \cite{transformer} where the inputs are split into smaller parts according to the input size of each head. Each part is treated by its own attention module and the outputs of these modules are concatenated. For spatial attention, we fix the number of heads at  $|J_1(t)|$, one for each dimension of the 3D coordinates.

\noindent \textbf{Self Temporal Motion Attention.}
For the self-temporal motion attention, we consider the entire skeleton and observe the  motion of its joints over time, i.e., we try to find the links between the position of the joints from one frame to another. This is performed in the same way as for self-spatial skeleton attention by using Eq.~\eqref{eq:temporal attention} and~Eq.~\eqref{eq:weights}. However, here $a_i{=}b_i{=}c_i$ represent the entire skeleton at time $t{=}i$, $dim{=}d$ and Q, K, and V are of size $dim {\times} T$. We also use the multi-head version of the attention, but here the number of heads can be set as a hyperparameter. 

\subsection{Motion Decoder}
The decoder receives the encoder's output $Z$ as well as the reaction sequence $Y$. It is composed of four attention modules as illustrated in Figure~\ref{fig:overview}. The self-attention modules work in the same way as the encoder but take $Y$ to which we add the positional encoding as input.\\
\noindent \textbf{Interaction Spatial Skeleton Attention.}
The interaction spatial skeleton attention module looks at the relations between the joints of the interacting skeletons at a given frame. The attention is also computed using Eq.~\eqref{eq:temporal attention} and~Eq.\eqref{eq:weights} but here the query matrix Q comes from the reaction sequence Y and the key and value matrices K and V come from the encoder output Z.

\noindent \textbf{Interaction Temporal Motion Attention.}
The interaction temporal motion attention module looks at the relations between the frames from the action sequence and the frames from the reaction sequence. Discovering these relations enable the synchrony of the generated reaction. Likewise, the query matrix Q comes from the reaction sequence Y but the key and value matrices K and V come from the encoder output Z.

In both the encoder and decoder, before each attention module, the input is normalized, and after each module, the output is also normalized and added to a residual connection of the non-normalized module input like in~\cite{transformer}. The spatial and temporal attentions are computed in parallel and are added after passing through all modules. This final output then goes through a feed-forward layer and is added to the residual connection.
The architectures described here for the encoder and the decoder correspond to a single layer of the encoder and one single layer of the decoder. There are $N{=}6$ of each of these layers, and the input of layer $h$ is the output of layer $h{-}1$. Finally, after the last decoder layer, the output goes through a final linear layer to get the reaction sequence.

\subsection{Skeleton Adjacency and Interaction Distance}


Recently many works using skeletons also use a graph representation which was proved to be a particularly efficient representation for action recognition~\cite{ZHU2021107920,PLIZZARI2021103219}. Building a graph for a skeleton is particularly intuitive in that the joints of the skeletons are already linked together by body segments. However, in our case, using a graph representation might be ill-fitted. Indeed while graphs provide information about the skeleton structure and help us concentrate on the most interesting parts of the skeleton, they would limit us when modeling the interaction. The information we have about the interaction is contained in the attention between the encoder and the decoder and the relations in the skeleton graph are very different from the relations between the joints of the two skeletons (all relations are possible). However, graphs can still provide important information that we can use to improve our generation.


\noindent \textbf{Skeleton Adjacency Module.}
We can use the information contained in the graph representation by looking at the adjacency matrices of the joints. We use three adjacency matrices that we combine to create a mask. 
The three matrices are based on the ones used by~\cite{PLIZZARI2021103219}: (i) the identity matrix $\mathbf{I}$ used to represent the joints themselves; (ii) the matrix of inward relations $\mathbf{In}$ which are the paths from the extremities (head, hands, and feet) to the root joint (torso or pelvis), and (iii) the matrix of outward relations $\mathbf{Out}$ which represents the paths from the root joint to the extremities. The three matrices of sizes  $|P_t| {\times} |P_t|$ are then added to get the mask matrix $\mathbf{M}{=}\mathbf{I}{+}\mathbf{In }{+}\mathbf{Out}$ that we apply to the attention matrix $\mathbf{Att}$ of size $|P_t| {\times} |P_t|$ to hide values that are not part of the graph as illustrated in the top right part of  Figure~\ref{fig:overview}.  
\begin{equation}
\label{eq:mask}
    \mathbf{Att}_{i,j}= 
\begin{cases}
    \mathbf{Att}_{i,j},                & \text{if } \mathbf{M}_{i,j}\neq 0\\
    0,                              & \text{if } \mathbf{M}_{i,j}= 0
\end{cases}
\end{equation}


\noindent \textbf{Interaction Distance Module.}
Interaction attention, which is also the attention between the encoder and the decoder, can also use a graph representation \cite{ZHU2021107920}, but this graph cannot be fixed since the interesting links between joints vary from class to class e.g., for ``punching" we are interested in the link between the hand and the head but not for ``kicking". Ultimately, it is the spatial attention between the encoder and decoder that discovers the important links between the two skeletons. However, as suggested by~\cite{ZHU2021107920}
we can add prior knowledge to the attention to help us model the interaction for some classes. This information is the distance between the joints of both skeletons, i.e., joints that are close to each other are more likely to interact than those that are far away:
\begin{equation}
\label{eq:interaction_distance}
\mathbf{Dist}_{i,j} = -\|J_{action}^{i}(t)-J_{reaction}^{j}(t)\|_2,
\end{equation}
where $J_{action}^{i}(t)$ and $J_{reaction}^{j}(t)$ are the joint $i$ and $j$ of the action and reaction skeletons at time $t$, $\mathbf{Dist}$ is a matrix of size $|P_t| {\times} |P_t|$. 
Unlike the graph for self-spatial attention, we do not use the distance matrix to create a mask because some of the relations between the two skeletons are not defined by the distance between the joints (e.g., waving and waving back), thus using the distance matrix as a mask would prevent such relations from being discovered. We add $\operatorname{softmax}(\mathbf{Dist})$ to the attention matrix to keep all the information that interests us, as illustrated in the top right part of Figure~\ref{fig:overview}. By using the softmax function on the distance matrix, we add values of the same order to the attention matrix while making shorter distances more important. 

\subsection{Objective Optimization}
We use two loss functions to direct our model. The first one is the sequence loss ($L_s$) which compares the generated sequence with the corresponding ground truth using the Mean Square Error (MSE) :
\begin{equation}
\label{eq:loss_skeleton1}
    L_s = \dfrac{1}{T}\dfrac{1}{k}\sum_{t=1}^{T}\sum_{i=1}^{k}(J_{i}(t)-\hat{J}_{i}(t))^2,
\end{equation}
where $J_{i}(t)$ is the position of the real joint $i$ at time $t$ and $\hat{J}_{i}(t)$ the position of the generated joint $i$ at time $t$. The second is the first frame loss ($L_{ff}$) used to add constraints on the first two frames by ensuring that the motion between the two is realistic and limits the discontinuities that can happen at the beginning of the sequences.  This loss is necessary as otherwise the model sometimes ignores the initial input frame and generates a sequence based on its own inferred initial position. For this loss, we also use the MSE but on the difference between the two first frames:
\begin{equation}
\label{eq:loss_skeleton2}
L_{ff} = \dfrac{1}{k}\sum_{i=1}^{k}((J_{i}(2)-J_{i}(1))-(\hat{J}_{i}(2)-\hat{J}_{i}(1)))^2.
\end{equation}


\subsection{Implementation Details}
We train our InterFormer using  Torch 1.8.1 on a PC with two 2.3Ghz processors, 64G RAM, and an Nvidia Quadro RTX 6000 GPU. We use the Adam  optimizer~\cite{adam} with $\alpha{=}0.0001$, $\beta_1{=}0.9$, $\beta_2{=}0.98$, and $\epsilon{=}1{\times}10^{-9}$. The  batch  sizes  are  set  to  128 for SBU and DuetDance and 64 for K3HI. 
InterFormer works even if we do not provide the original position of the reaction sequence (the first frame of the sequence) as input, 
but this can cause the generator to produce a skeleton very far from its actual location, which will lead to a bad generation. To solve this during testing, we give as input to the decoder the first frame of the sequence which gives information about the original location of the skeleton. During testing, we generate sequences of variable lengths depending on the length of the input action motion. The sequences are generated in an auto-regressive manner and the model generates an end-of-sequence value to indicate the end of the motion generation. If the motion is generated correctly, then this value will correspond to the end of the input action sequence.

\section{Experiments}
We conducted comprehensive experiments to evaluate our proposed approach by comparing state-of-the-art models on three datasets. We also visualize the ability of action-reaction generation. Finally, we perform ablation studies to evaluate the effectiveness of using spatial attention and our skeleton adjacency and interaction distance modules.

\subsection{Datasets}
\noindent \textbf{SBU Dataset} \cite{kiwon_hau3d12} contains 8 classes of simple interaction motions: walking toward, walking away, kicking, pushing, shaking hands, hugging, exchanging, and punching. The data which are too noisy, and in particular the class ``hugging", have been removed from this dataset.  
The ``walking away" and ``walking toward" classes have the same reactions (standing still), so we decided to fuse those two classes into a single ``walking" class. This leaves us with 6 classes, 195 training, and 30 test samples. 

\noindent \textbf{K3HI Dataset} \cite{K3HI} contains the same  8 classes as SBU aside from the ``hugging" class which is replaced by ``pointing". Also, unlike SBU, ``approaching" and ``departing" have reactions that are different, so we do not fuse the two classes. We also removed the noisy samples from the dataset but this time, we normalize the data in the same way as SBU was normalized by the authors. This leaves us with 236 training samples and 28 test samples.

\noindent \textbf{DuetDance Dataset} \cite{duetdance} contains 5 classes of dance motions: cha-cha, jive, rumba, salsa, and samba. Given the nature of the dataset, the motions are more complex than those in SBU and K3HI, and there are a lot of intra-class variabilities. We do not perform normalization, but since most samples are very long sequences (up to 160s), we decided to cut each sequence into smaller sequences of 50 frames (2s), leading to 273 training samples and 3991 test samples.

For all three datasets, the poses are represented by their absolute 3D coordinates, furthermore, training and testing splits are selected randomly for fair comparisons. Duet-Dance was provided with neither train/test split nor subject information, and we used a random split. For the two others, the evaluation proposed by their respective authors is made using k-fold validation so we decided to split the dataset between train and test, randomly for K3HI and by selecting all the samples from a random subject for SBU.

\subsection{Evaluation Metrics}
We use metrics commonly used in motion generation. Metrics used for motion prediction based on the distance between the generated sample and the ground truth are not fit for reaction generation as several different motions can be considered good reactions to the same action. While this choice of metric can seem contradictory with our losses that use direct comparison with the ground truth, it is important to understand that our evaluation metrics do not contain direct information about the skeleton that our network is supposed to generate and could not be efficiently used as losses.

\noindent \textbf{Classification Accuracy} measures how well our generated samples are classified by a motion classifier. We use the DeepGRU classifier~\cite{DeepGRU}. We only train and test the classifier on the reaction part of the interaction, so the results are not influenced by the action, which is always the ground truth. We report the percentage of correctly classified samples for each class and the average over the entire test set.

\noindent \textbf{Fréchet Video Distance} (FVD) is an adaptation of the Fréchet Inception distance (FID) \cite{FID} for video sequences \cite{FVD}. FVD computes the distance between the ground truth and the generated data distribution.
\begin{equation}
\label{FVD}
\text{FVD} = \left|\mu_{gt} - \mu_{gen}\right|^2 + \text{tr}\left[\mathbf{C}_{gt} {+} \mathbf{C}_{gen} {-} 2\left(\mathbf{C}_{gt} * \mathbf{C}_{gen}\right)^{1/2}\right],
\end{equation}
where $\mu_{gt}$, $\mu_{gen}$ and $\mathbf{C}_{gt}$ and $\mathbf{C}_{gen}$ are the means and covariance matrices of the deep features from ground truth and the generated samples respectively, tr($\cdot$) is the trace. The deep features are obtained from the classifier used for the classification accuracy

\noindent \textbf{Diversity Score.} Following the metric defined by  \cite{dancingmusic,zhang2018perceptual} we compute the average deep feature distance between all the samples generated by each method and then compare it to the average deep feature distance of the ground truth. 
A low diversity score means that the generated samples have a diversity close to that of the ground truth and a high score means that the diversity is either lower (all motions are more similar) or higher (more noise in the generation). The average deep feature distance is calculated as follows:
\begin{equation}
    \label{diversity}
    div = \dfrac{1}{b(b-1)}\sum_{i=1}^{b}\sum_{j=1}^{b}||F_i-F_j||_2,
\end{equation}
where $b$ is the number of samples considered, $F_i$ and $F_j$ are deep features of the samples $i$ and $j$, respectively. The score is obtained using $div_{gt}$ the diversity distance of the ground truth and $div_{gen}$ the diversity of the generated samples.
\begin{equation}
    \label{diversity score}
    score = 100\times\dfrac{|div_{gt}-div_{gen}|}{div_{gt}}.
\end{equation}

\subsection{Baselines}
To our knowledge, there is no work that deals with the generation of the reaction to an action, so to be able to compare our results to others from the literature, we employ a method for human interaction generation and a method for human motion prediction to show methods used on a range of applications.

\noindent \textbf{Zero Velocity baseline} (ZeroV) \cite{martinez_human_2017} is a simple baseline where all generated frames are the same (in our case the initial pose), there is no motion for this baseline. Using ZeroV as a comparison is useful to see what the quantitative result of an obviously bad method are like and help see if the results from the other methods are actually good. We do not show the results for ZeroV in our qualitative evaluation as they are uninteresting since no motion is produced. We do not use them in our user study for the same reason. 

\noindent \textbf{Multimodal Variational Recurrent Neural Network} (VRNN) \cite{multimodal_interaction}  deals with the prediction of the future frames of a two-person interaction based on a historical sequence using  variational RNNs. The next frame of the reaction is predicted using the past frames of the reaction and information on the past frames of the action; the action is predicted in the same way using the information on the reaction. The past frames are the historical sequence at the beginning and later in the sequence the generated frames. 
We modified the network to fit our problem. Originally the network takes $n$ historical frames for both action and reaction as input and generates $m$ frames for both action and reaction. We modify some parameters so the network takes $n+m$ frames for the action but only $1$ for the reaction and we generate $n+m-1$ frames of reaction motion. Otherwise, we use the default settings provided by the author for the hyper-parameters.

\noindent \textbf{Mix-and-Match Perturbation} (MixMatch) \cite{mix-and-match-perturbation}  uses a  recurrent encoder-decoder network with a conditional variational autoencoder block to predict the motion of a single person based on a historical sequence. 
However, the authors present their method as a general prediction method and the code they provide uses the first half of an image to predict the second half. Since the specific code used for motion prediction is not available we use the one provided by the author but with 3D skeletons data and with the values of the hyperparameters mentioned in \cite{mix-and-match-perturbation} for human motion prediction. To ensure a fair comparison we need to base the generation of the reaction on an initial frame but directly using 3D coordinates led to strong discontinuities between the initial position and the generation. To solve this and make the comparison fairer we work with the speed of the motion that we then apply to the skeleton corresponding to the initial position.

\noindent \textbf{Progressively Generating Better Initial Guesses} (PGBIG) \cite{Ma_2022_CVPR} is an architecture that uses Spatial Dense Graph Convolutional Networks and Temporal Dense Graph Convolutional Networks alternatively to extract spatio-temporal feature and predict human motion. We use the code provided by the authors unchanged and with the recommended parameters. We give the action motion followed by the first frame as input and predict the reaction motion.

\noindent \textbf{Spatio-temporal Transformer} (STT) \cite{aksan2021spatio} is a Transformer based architecture that uses attention to find temporal and spatial correlations to predict human motion. As for PGBIG, we use the code provided by the authors without changes and with the recommended parameters. As input, we use the action motion followed by the first frame and predict the reaction motion.

\begin{table*}[!t] \small
	\centering
		\caption{\textbf{Left}: Classification accuracy for each class of the SBU, DuetDance, and K3HI datasets. \textbf{Right}: User study for each class of the SBU, DuetDance, and K3HI datasets.}
		\resizebox{1\linewidth}{!}{%
\begin{tabular}{@{}c| c c c c c c c| c c c c c@{} } 
\toprule
Method & GT  & ZeroV \cite{martinez_human_2017} & VRNN \cite{multimodal_interaction} & MixMatch \cite{mix-and-match-perturbation} &STT \cite{aksan2021spatio}& PGBIG\cite{Ma_2022_CVPR} & InterFormer & GT  & VRNN~\cite{multimodal_interaction} & MixMatch~\cite{mix-and-match-perturbation} & InterFormer \\
\midrule
\multicolumn{8}{c|}{Classification Accuracy $\uparrow$}&\multicolumn{4}{c}{User Preference $\uparrow$}\\ \hline
 \multicolumn{12}{c}{SBU}\\ \hline
Walking         & 100.0 & 0.0       & 58.3              & \textbf{100.0}    &91.7   &58.3   &\textbf{100.0}    &  34.2\% & 21.4\% & 15.5\% & 28.9\%\\
Kicking         & 66.7  &  66.7     & 0.0               & 0.0               &0.0    &0.0    &\textbf{33.3}      &  38.8\% & 23.8\% & 5.6\% & 31.8\%\\
Pushing         & 80.0  &  0.0      & \textbf{60.0}     & 0.0               &0.0    &0.0    &\textbf{60.0}      &35.6\% & 19.7\% & 15.4\% & 29.3\%\\
Shaking Hands   & 100.0 &  0.0      & 0.0               & 0.0               &0.0    &0.0    &\textbf{ 100.0}	&37.5\% &	21.8\% & 7.8\% & 32.9\%\\
Exchanging      & 80.0  &  0.0      & \textbf{80.0}     & 0.0               &50.0   &60.0   & 60.0              &  41.9\% & 19.4\% & 13.0\% & 25.7\%\\
Punching        & 100.0 &  33.3     & 0.0               & 33.3              &0.0    &0.0    &\textbf{100.0}     & 43.1\% & 19.3\% & 11.3\% &  26.3\% \\  \hline
Average         & 90.0  &  10.0     & 46.7              & 43.3              &40.0    &33.3   &\textbf{80.0}      & 38.5\% & 20.9\% & 11.4\% & 29.2\% \\ \hhline{= = = = = = = = = = = =}
\multicolumn{12}{c}{DuetDance}\\ \hline
Cha-Cha     & 28.0  & 1.8   & 26.4          & 19.2          &\textbf{37.1}   &28.6   & 26.7 & 45.9\% & 17.8\% & 5.5\% & 30.8\%\\
Jive        & 24.6  & 0.4   & 13.8          & \textbf{25.8} &16.7   &19.7   & 22.8          & 48.4\% & 13.2\% & 6.7\% & 31.7\% \\
Rumba       & 34.8  & 0.7   & \textbf{36.4} & 30.0          &30.0   &34.5   & 32.0          & 40.7\% & 16.9\% & 8.2\% & 34.2\% \\
Salsa       & 27.8  & 93.1  & \textbf{29.5} & 28.9          &10.0   &20.2   & 28.1	        & 49.3\% & 12.8\% & 7.1\% & 30.8\%	\\
Samba       & 22.2  & 18.6  & 21.0          & 17.2          &18.2   &\textbf{24.4}   &\textbf{24.4}  & 44.5\% & 15.8\% & 6.3\% & 33.6\% \\ \hline
Average     & 28.0  & 24.6  & 26.2          & 24.9          &24.4   &25.7   &\textbf{27.1}  & 45.8\% & 15.3\% & 6.7\% & 32.2\% \\ \hhline{= = = = = = = = = = = =}
\multicolumn{12}{c}{K3HI}\\ \hline
Approaching & 100.0 & 25.0  & \textbf{75.0}     & 50.0          &50.0   &0.0            & 0.0           & 34.9\% & 23.1\% & 14.1\% & 27.9\% \\
Departing   & 33.3  & 33.3  & 0.0               & \textbf{33.3} &\textbf{33.3}   &\textbf{33.3}           &\textbf{33.3}  & 34.2\% & 24.2\% & 13.2\% & 28.4\% \\
Kicking     & 40.0  & 80.0  & 0.0               & 40.0          &40.0   &0.0            &\textbf{60.0}  & 31.8\% & 21.7\% & 18.8\% & 27.7\% \\
Pushing     & 100.0 & 33.3  & 33.3              & 33.3          &33.3   &33.3           &\textbf{66.7}  & 33.1\% & 24.8\% & 13.5\% & 28.6\% \\
Shaking     & 50.0  & 0.0   & 50.0     & 50.0 &\textbf{100.0}  &\textbf{100.0}          & 50.0	& 36.9\% & 21.4\% & 10.8\% & 30.9\% \\
Exchanging  & 0.0   & 0.0   & 0.0               & 0.0           &0.0    &0.0            &0.0            & 36.3\% & 20.5\% & 13.9\% & 29.3\% \\
Punching    & 100.0 & 25.0  & \textbf{50.0}     &25.0           &0.0    &\textbf{50.0}           & \textbf{50.0} & 33.9\% & 21.7\% & 16.9\% & 27.5\% \\
Pointing    & 100.0 & 50.0  & \textbf{100.0}    & 0.0           &25.0    &25.0           &\textbf{100.0} & 37.3\% & 20.2\% & 16.1\% & 26.4\% \\ \hline
Average     & 67.9  & 35.7  & 39.3              & 28.6          &32.1   &25.0           &\textbf{46.4}  & 34.8\% & 22.2\% & 14.7\% & 28.3\% \\
 \bottomrule
 \end{tabular}}
	\label{tab:quantitative}
\end{table*}


\subsection{State-of-the-Art Comparisons}


\begin{figure}[!t]
    \centering
    \includegraphics[width=1\linewidth]{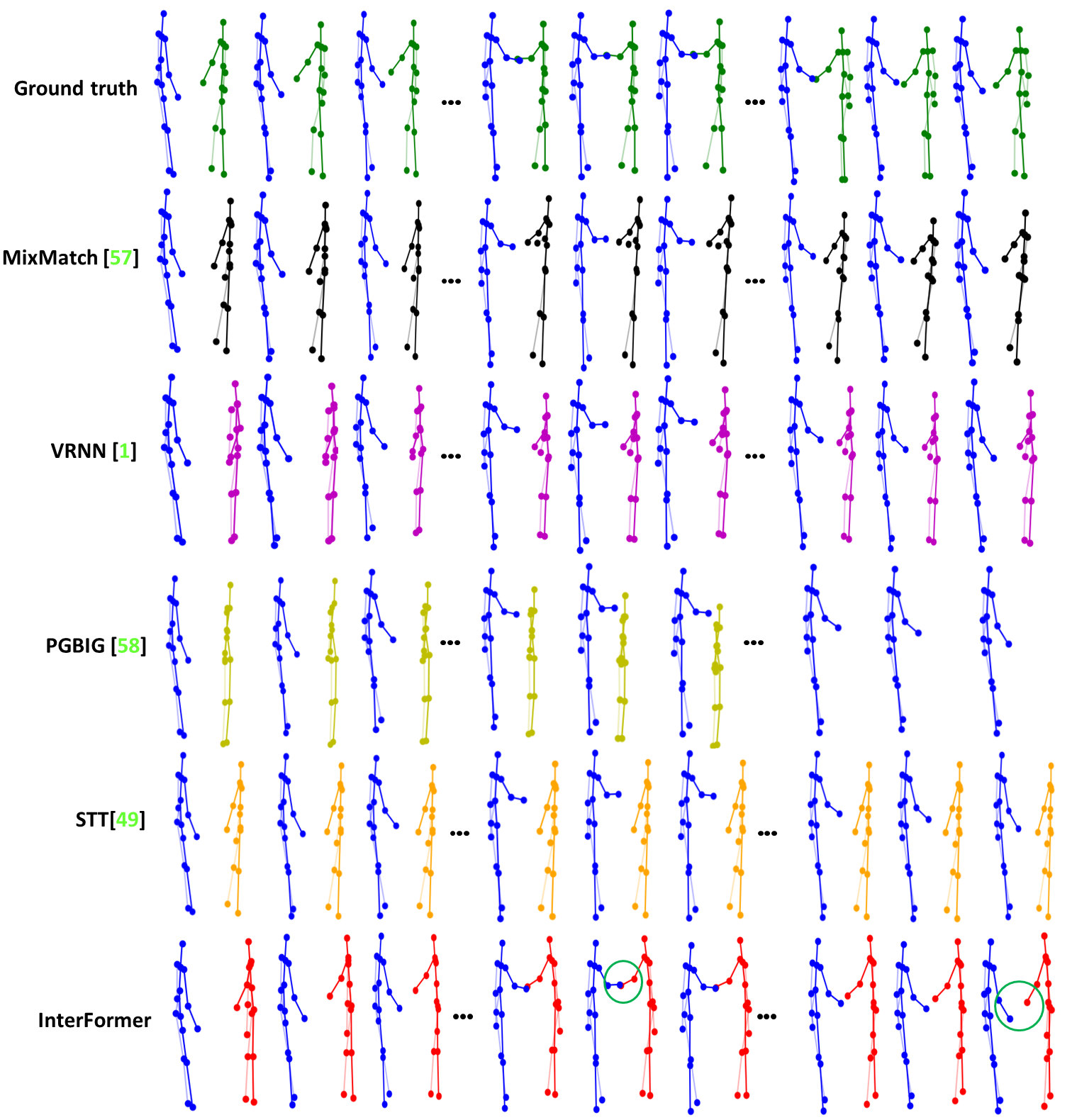}
    \caption{\textbf{Qualitative results.} In blue the action motion is used as a condition. In other colors, the reaction is either from the ground truth or generated by the different models. Shaking hands class from the SBU dataset.}
    \label{fig:intro_SBU}
\end{figure}

\begin{figure*}[!t]
    \centering
    \includegraphics[width=1\linewidth]{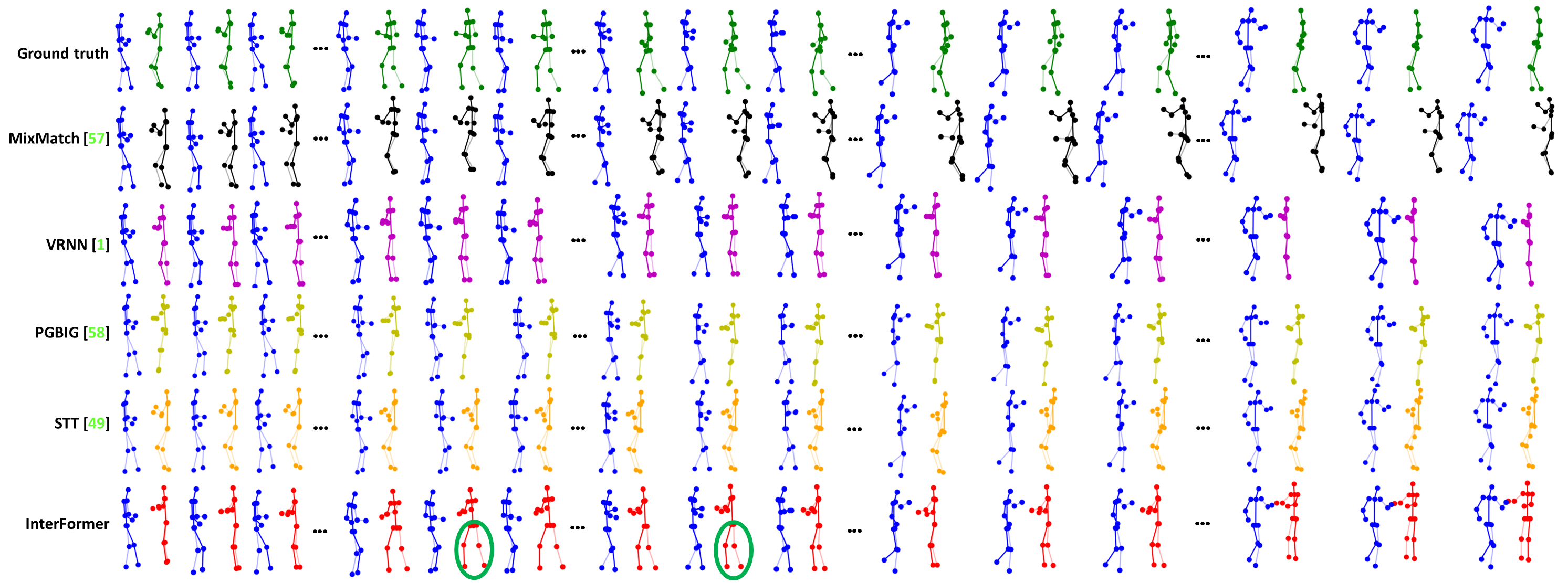}
    \caption{\textbf{Qualitative results.} In blue the action motion is used as a condition. In other colors, the reaction is either from the ground truth or generated by the different models. Cha-cha class from the DuetDance dataset.}
    \label{fig:qualtitative_DD}
\end{figure*}

\begin{figure}[!t]
    \centering
    \includegraphics[width=1\linewidth]{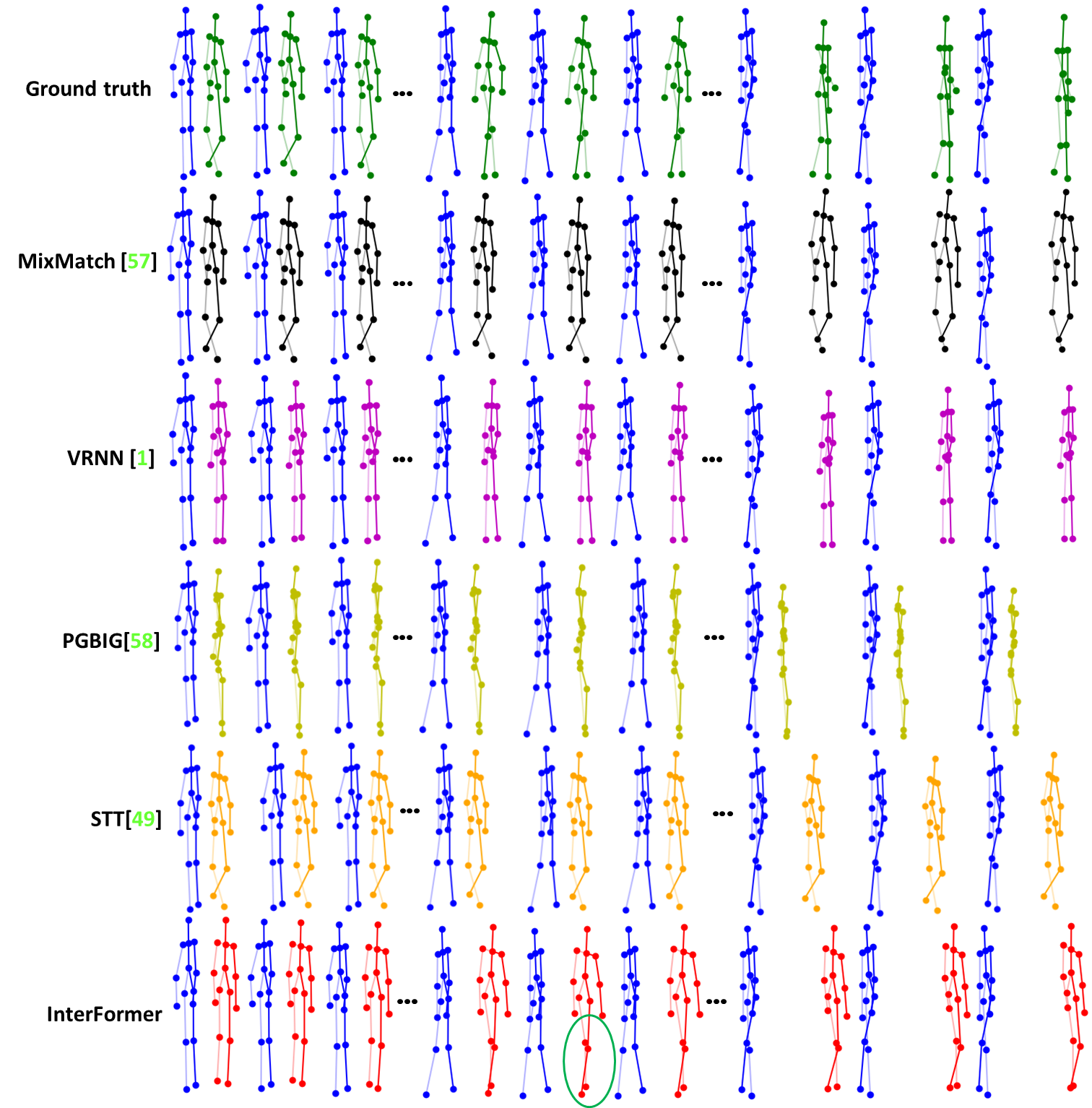}
    \caption{\textbf{Qualitative results.} In blue the action motion is used as a condition. In other colors, the reaction is either from the ground truth or generated by the different models. Departing class from the K3HI dataset.}
    \label{fig:qualtitative_K3HI}
\end{figure}


All presented evaluations were obtained on a model trained on the considered dataset. This is true for our Interformer as well as the baselines.

\noindent \textbf{Quantitative Evaluation.}
Table~\ref{tab:quantitative} (left) shows the classification accuracy for SBU, DuetDance, and K3HI. Our method outperforms the five others on all the datasets. For SBU, we obtain results very close to the ground truth, and we outperform the other methods on all classes but ``exchanging'' where \cite{multimodal_interaction} get better results and vastly outperform the simple ZeroV baseline. InterFormer is able to generate simple motions that are realistic enough to be correctly classified. We can see however that on ``kicking'' we score less than ZeroV, this is due to the small size of the SBU dataset. A few misclassifications will cause a sharp drop in classification accuracy, and as we can see, ``Kicking'' is the class that has the lowest accuracy on the ground truth as the reaction can be similar to those of punching and pushing. The good performance of ZeroV in some classes can be explained by the fact that the overall accuracy is below chance (16.7\%). This means that the classifier is unable to properly classify the motion from ZeroV as it only shows unmoving skeletons and for some classes, the two skeletons start in a neutral position that carries no information about the action. All these cause the classifier to fail at classifying the sample and likely classify many samples as ``kicking'', including some that are from the ``kicking'' class leading to the high score in this class.

For K3HI, we can see that the results are worse than for SBU for all methods and even for the ground truth. This is due to the very noisy nature of the K3HI dataset even after removing the worse samples (that showed extreme deformation and no recognizable motion), the exchanging class has a 0\% recognition rate even for the ground truth. However, our method provides better results than the two others in all classes except ``approaching'' which may be due to the noisy nature of the data for this class. VRNN obtaining very high results in this class might be a consequence of the wrong classification present in many of the classes (a lot of samples are classified as approaching). For shaking PGBIG and STT obtain better results but since the results are worse overall quantitatively and qualitatively this can be explained by the classifier  putting many samples in that class as we have explained for ZeroV.

For DuetDance, the classification accuracy for all methods and the GT is much closer than for the other datasets. This is due to the complex motions contained in the dataset with a lot of intra-class variabilities. Furthermore, we use sequences of 50 frames which are short enough that some sequences from two different classes can be very similar. We can still notice that our method provides results that are the closest to the ground truth and that, unlike the five other methods no class has a score below chance (i.e., 20\%) which means that our results are more consistent and closer to the ground truth, despite being beaten on some individual class e.g., STT score 37.1\% on ``cha-cha'' but only 10.0\% on ``salsa'' while we score 26.7\% and 28.1\%, respectively.

In Table~\ref{tab:quanti} we show the FVD and diversity score for all methods on all datasets. We outperform VRNN, MixMatch STT and PGBIG on the FVD measure, often by a large margin meaning that the features extracted by the classifier are closer to the features of the ground truth than for \cite{multimodal_interaction} and \cite{mix-and-match-perturbation}. For the diversity score, we also outperform the two other methods and provide diversity that is close to that of the ground truth. We can see a significant increase in K3HI. This is due to the noisy nature of the dataset, which means that the diversity distance of the ground truth takes into account the noise of the sample, we, however, manage to score the closest to the diversity of the ground truth when compared to the other methods, without generating noisy samples. This can also explain why PGBIG diversity is better than ours despite performing much worse in terms of classification and qualitative results.

\begin{table}[!t] \small
	\centering
		\caption{FVD and diversity on all datasets. }
		\resizebox{1\linewidth}{!}{%
\begin{tabular}{@{}c| c c c c c c@{} } 
\toprule
\multirow{2}*{Method} & \multicolumn{3}{c}{FVD $\downarrow$} & \multicolumn{3}{c}{Diversity $\downarrow$} \\ \cmidrule(lr){2-4} \cmidrule(lr){5-7}  
 & SBU & DuetDance & K3HI & SBU & DuetDance & K3HI \\
\midrule
ZeroV \cite{martinez_human_2017} &493.3&41058.1  &392.1&65.1&47.2  &19.3\\
VRNN \cite{multimodal_interaction}           & 113.61 & 789.23   &195.47    &11.5   &6.1    &16.8 \\          
MixMatch \cite{mix-and-match-perturbation}   & 314.38 & 1460.44  &406.63     &45.3   &0.9    &32.2 \\          
STT\cite{aksan2021spatio}   &321.04  & 2610.95       &7579.87      &47.8       &3.9 &27.6 \\
PGBIG\cite{Ma_2022_CVPR}  &267.27  & 317.0       &379.4      &35.7       &1.5 &\textbf{10.1} \\ 
InterFormer (Ours)                          & \textbf{48.78}  & \textbf{31.81}    & \textbf{125.40}    &\textbf{0.9}    &\textbf{0.4}    &13.7 \\          
\bottomrule
\end{tabular}}
\label{tab:quanti}

\end{table}

\noindent \textbf{User Study.} 
To evaluate the quality of the generated videos, we also conduct a user study. Specifically, the users are given four videos (two generated by existing methods VRNN and MixMatch, one generated by our proposed InterFormer, and one real video) with the corresponding class label. Each participant needs to answer one question: `Which video is more realistic regardless of the input label?'. 20 users have unlimited time to select their choices. PGBIG and STT are not represented in this study due to the extremely low quality of the results, as illustrated by our qualitative results.
The results are shown in Table~\ref{tab:quantitative} (right). We can see that the users show more preference for our method than the other two methods, which indicates the results generated by ours are more realistic.

\begin{figure*}[!t]
    \centering
    \includegraphics[width=1\linewidth]{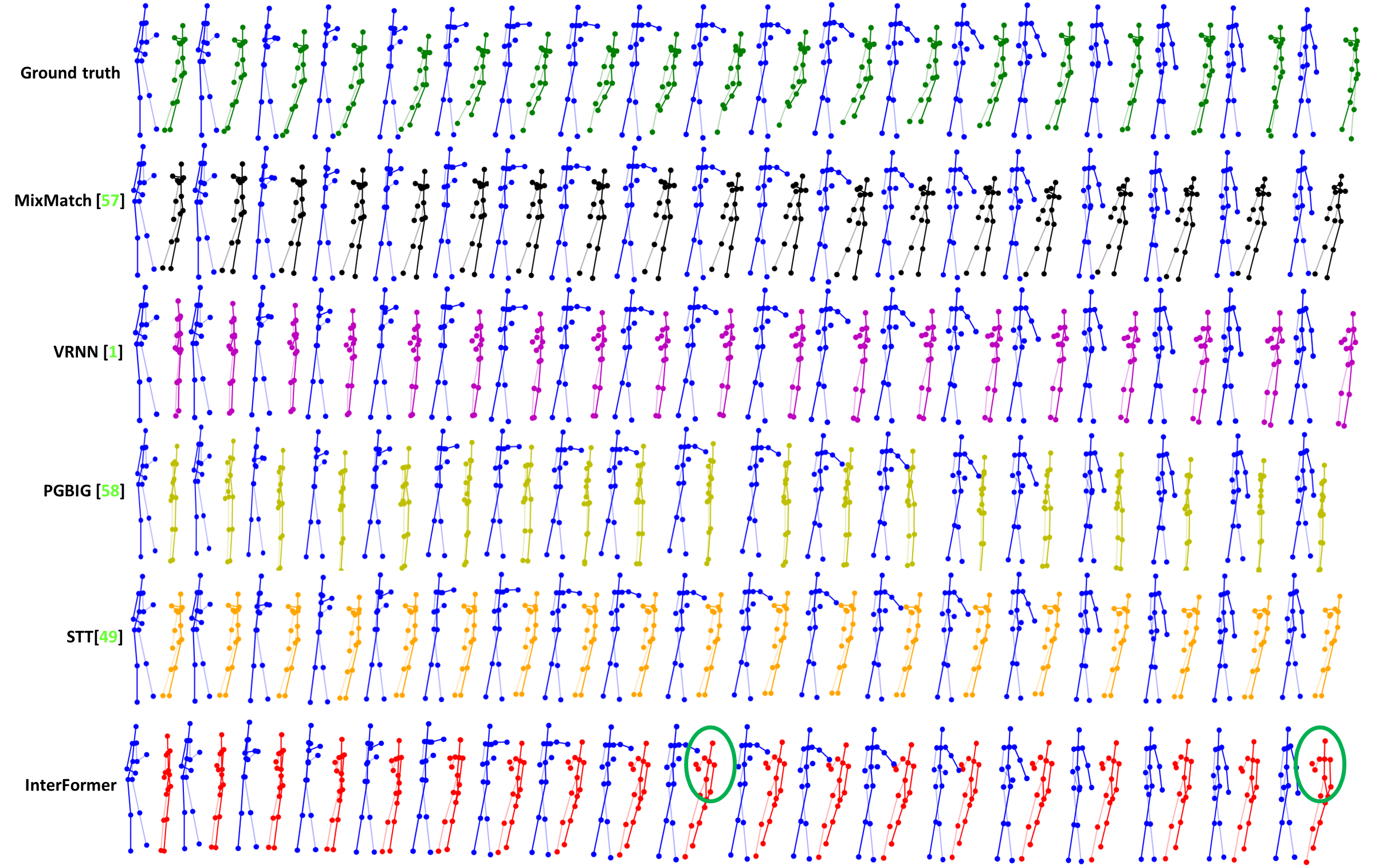}
    \caption{\textbf{Qualitative results.} In blue the action motion is used as a condition, in other colors, the reaction is either from the ground truth or generated by the different models. Punching class from the SBU dataset.}
    \label{fig:qualtitative_SBU}
\end{figure*}

\noindent \textbf{Qualitative Evaluation.}
We show in Figures~\ref{fig:intro_SBU}, \ref{fig:qualtitative_DD}, \ref{fig:qualtitative_K3HI}, and  \ref{fig:qualtitative_SBU}
visualizations of the generated sequences on the SBU (two sequences)  DuetDance and K3HI datasets respectively. We show from top to bottom: the ground truth, results for \cite{mix-and-match-perturbation}, results from \cite{multimodal_interaction}, results from \cite{Ma_2022_CVPR}, results from \cite{aksan2021spatio} and results from our InterFormer. In blue is the action motion, which serves as a condition and is in all cases the ground truth. Green, black, magenta, yellow, orange, and red are the reactions for the GT and the five methods. More visualizations, as well as animations, are available in our supplementary materials. 

In Figure~\ref{fig:intro_SBU}, we show an interaction from the ``shaking hands" class of SBU. It shows that our method is able to generate the motion better than the two other methods. For \cite{mix-and-match-perturbation}, the character raises its hand to shake but never comes really close to the other character's hand and also shifts its entire body backward toward the end of the sequence. \cite{multimodal_interaction} generates a motion that raises slightly the hand but is then stuck in this position. \cite{Ma_2022_CVPR} does not generate a shaking hand motion and fails to generate poses for the entire length of the action. STT \cite{aksan2021spatio} also fails to generate a shaking hand motion. Our method generates motion that is very close to the ground truth and contains the three main steps of the motion: raising the hand, shaking, and going back to starting position.
Figure~\ref{fig:qualtitative_SBU} shows a sample from the ``punching" class from the SBU dataset. We see that we generate a better motion even if there are differences with the ground truth. The character is pushed to the side by the punch and then comes back to a normal position at the end of the sequence. The two other methods also generate a reaction to the punch, \cite{multimodal_interaction} moves slightly backward, and \cite{mix-and-match-perturbation} moves its upper body to avoid the punch. \cite{Ma_2022_CVPR} does not generate a motion that looks like a reaction to the punch and presents noise with the vertical position of the skeleton suddenly changing from one frame to the other. \cite{aksan2021spatio} generates a slight motion of  being pushed back but the motion continues without trying to go back into a neutral position. It seems, however, that the upper body also became smaller during this motion. The two methods also stay in this avoiding pose and do not go back to a more normal position.
In Figure~\ref{fig:qualtitative_DD} we show a sample of the ``cha-cha" class from the challenging DuetDance dataset. We can see that \cite{mix-and-match-perturbation} produce a motion that resembles a dance even if different from the ground truth, however as the action character moves backward (better seen in the animated sequence in our supplementary material), the generated reaction stays in place, and the distance between both characters grows over time. With \cite{multimodal_interaction}, the distance between the two characters does not grow, but there is barely any motion for the entire sequence.  In motion, it looks like the reaction character is gliding toward the action character (better seen in the animation in our supplementary material). Here \cite{Ma_2022_CVPR} and \cite{aksan2021spatio} generate something close to \cite{multimodal_interaction} with little motion, but the distance between the two skeletons does not grow. \cite{aksan2021spatio} also present deformations in the arms. 
Our method is able to generate a motion that stays close to the ground truth and follows the action character in space without gliding like \cite{multimodal_interaction} this can be seen by the change of  position of the legs across the sequence. It is only toward the end that the motion differs from the ground truth and even then, the motion still resembles dancing. 

In Figure~\ref{fig:qualtitative_K3HI}, we see a sample of the departing class from the K3HI dataset. It shows both characters walking away from each other. This behavior is always reproduced in the samples generated by the three methods, but \cite{multimodal_interaction} does not show much motion and simply glides away while~\cite{mix-and-match-perturbation} shows more motion of the legs but keeps the noise present in the first frame during the entire sequence. Once again \cite{Ma_2022_CVPR} does not generate a proper motion, and this time it shows deformation in the skeleton that stays for the entire duration of the motion. Likewise, \cite{aksan2021spatio} is unable to generate a proper walking motion.
Our method, on the other hand, generates a realistic walking motion with both arms and legs moving to move apart from the first character.

The very poor performances  of PGBIG \cite{Ma_2022_CVPR} and STT \cite{aksan2021spatio}, our two baselines with unmodified code, can be explained by the fact that they were designed for human motion prediction. With human motion prediction, we seek to reduce as much as possible the discontinuities between the input and the output while we want to generate a different skeleton to the one used as input which implies a very strong discontinuity. Also, methods for human motion prediction are typically trained to always take the motion of the same duration as input and predict sequences that always have the same length e.g., the input of $500ms$ to predict $1s$ of motion. With reaction generation, the length of the sequences can vary (greatly in the case of K3HI) and the unmodified motion prediction method might struggle with the varying lengths. This is illustrated by the early stop in the generation of \cite{Ma_2022_CVPR} in Figure~\ref{fig:intro_SBU} but also by the fact that \cite{aksan2021spatio} is unable to stop generating until it reaches the maximum sequence length of the dataset (not pictured in our figures).

\begin{figure}[!t]
    \centering
    \includegraphics[width=1\linewidth]{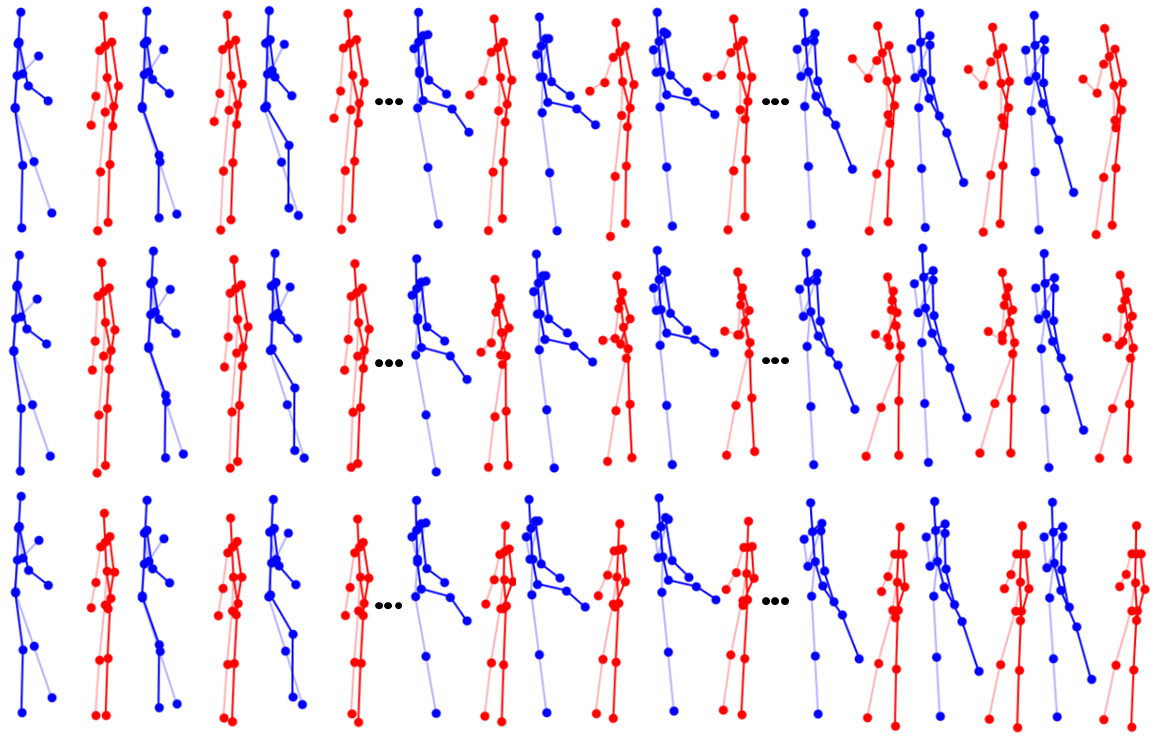}
    \caption{\textbf{Multi-modality results on SBU kicking class with noise.} We show three different motions generated by our Interformer based on the same input motion.}
    \label{fig:multimodality}
\end{figure}

\begin{figure}[!t]
    \centering
    \includegraphics[width=1\linewidth]{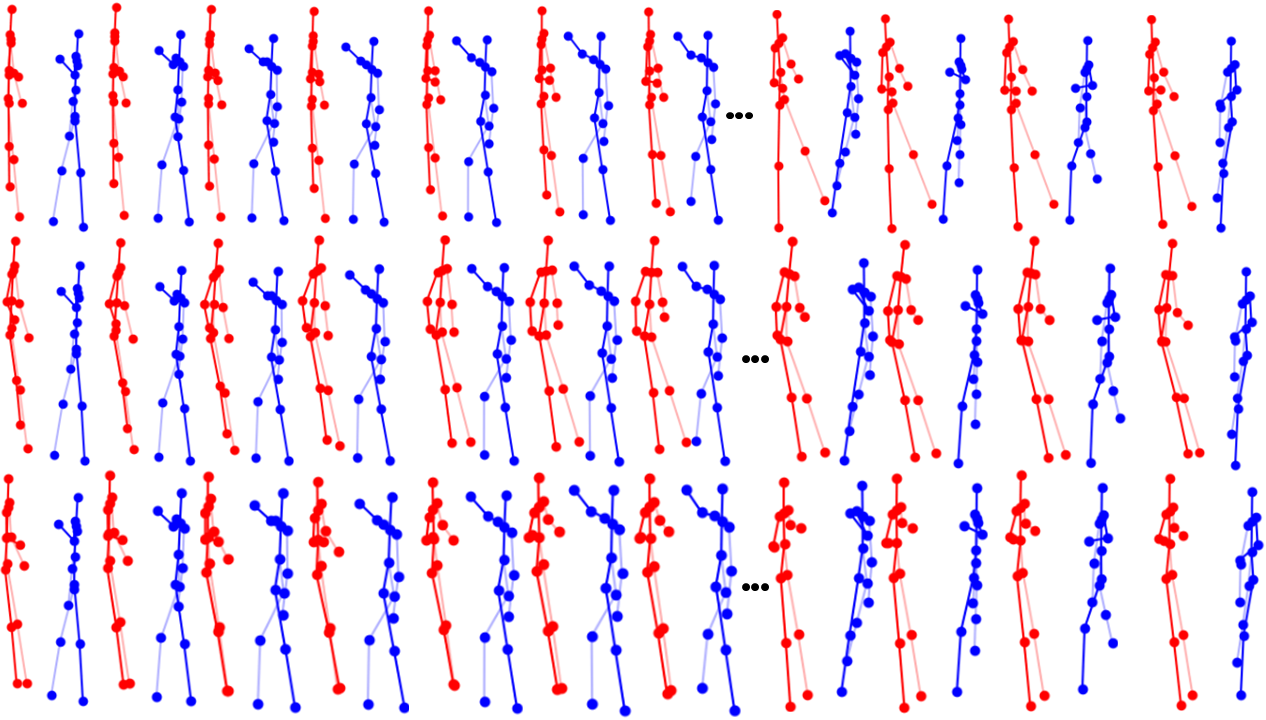}
    \caption{\textbf{Multi-modality results on SBU Punching class with noise.} We show three different motions generated by our Interformer based on the same input motion.}
    \label{fig:multimodality_2}
\end{figure}

\noindent\textbf{Multi-Modality Generation.}
The main issue with Transformer models is that their output is deterministic. To counter this we can add noise to the encoder input before the first feed-forward layer. This allows us to generate diverse outputs for the same input motion. We show in Figure~\ref{fig:multimodality} and Figure~\ref{fig:multimodality_2}  the ability of our method to generate diverse motions with a single input when adding noise in the encoder.

\subsection{Ablation Study}
To validate the effectiveness of each proposed component, we report the ablation studies on SBU with classification accuracy and diversity. 

\begin{table}
\centering
		\caption{ Ablation study of  Interformer on the SBU dataset.}
\begin{tabular}{@{}c| l c c @{} } 
\toprule
& Setup & Accuracy $\uparrow$ & Diversity $\downarrow$ \\
\midrule
S1 & Transformer                     & 53.3 &9.5\\ 
S2 &  S1 + Spatial Attention            & 66.7 &3.9\\ 
S3 & S2 + Skeleton Adjacency                    & 73.3 &1.7\\ 
S4 & S3 + Interaction Distance & \textbf{80.0} &\textbf{0.9}\\ 
 \bottomrule
 \end{tabular}
	\label{tab:abla}
\end{table}

\noindent\textbf{Ablation Models.} Our Interformer has four versions (i.e., S1, S2, S3, S4) as shown in Table~\ref{tab:abla}. (i) S1 means only using the original NPL Transformer network from \cite{transformer} modified to take as input and generate skeletons without any of our improvements. (ii) S2 adds to the global Transformer the spatial attention modules (self-spatial attention and interaction spatial attention). (iii) S3 adds the skeleton adjacency module to the self-spatial attention.
(iv) S4 is the full model and includes both the skeleton adjacency module and the interaction distance module.

\noindent\textbf{Effect of Spatial Attention.} We validate the effect of spatial attention, as shown in Table ~\ref{tab:abla}. Introducing the spatial attention results in significant improvement in classification accuracy by 13\% and diversity by 5.6, which means we improve the quality of the action-reaction sequences.

\noindent \textbf{Effect of Skeleton Adjacency.}  Using  a skeleton adjacency graph on attention improves the classification accuracy and diversity  by 7\% and 2.2, respectively. This improvement means that the model learns better relations between the different joints inside a skeleton. 



\noindent \textbf{Effect of Interaction Distance.} 
By adding the interaction distance module, we increase the results obtained by the skeleton adjacency module by 7\% on classification and 0.8 on diversity. These results show that the interaction distance module is able to help spatial interaction attention find the most interesting relations between the two skeletons and thus help generate better motions.

\begin{table}

\centering
		\caption{ Ablation study of  Interformer on the K3HI dataset}.
\begin{tabular}{@{}c| l c c @{} } 
\toprule
& Setup & Accuracy $\uparrow$ & Diversity $\downarrow$ \\
\midrule
S1 & Transformer                    & 14.3 &32.9\\ 
S2 &  S1 + Spatial Attention            & 28.6 &16.7\\ 
S3 & S2 + Skeleton Adjacency                   & 42.9 &\textbf{9.3}\\ 
S4 & S3 + Interaction Distance & \textbf{46.4} &13.7\\ 
 \bottomrule
 \end{tabular}
	\label{tab:abla_K3HI}

\end{table}

\noindent \textbf{Abaltion on K3HI.} Table \ref{tab:abla_K3HI} shows the ablation for the K3HI dataset and confirm our finding from the SBU ablation. The only difference is a lower diversity when using the graphs but not the interaction distance. We believe this to be due to the more noisy nature of the K3HI dataset, which deteriorates the diversity measures.

\noindent\textbf{Effect of Loss on The First Frames.} 
If we remove the loss on the first frames that allow us to keep a good coherency between the input initial position and the generation, we see a decrease in the generation quality: -3.3\% in classification accuracy and -5.2 in diversity score when compared to S4. When the input initial position is not properly taken into account the generated reaction skeleton can be far from the action skeleton. In SBU, for all action classes, the interactions consist of two persons close to each other. Since the model is not trained with samples where people are far from each other when we try to generate the reaction motion of a skeleton far from the action skeleton, little to no motion is generated. This explains the increase in performance brought by the use of the first frame loss.

\noindent\textbf{Effect of Multihead attention.}
Our Interformer uses the multihead version of attention for both temporal and spatial attention. These choices were made following results from the original Transformer network \cite{transformer} and our experiments which we report in Table \ref{ablationmultihead}. The results are obtained by modifying the number of heads for the different attention modules on the full Interformer model (S4 from the main paper ablation study). These experiments show that using the multihead temporal attention (T multihead) increases the classification accuracy by 10\%  and diversity by 8.6. By using only the spatial multihead attention (S multihead) we increase the diversity by 5.5. Using the multihead attention for both spatial and temporal attention led to an increase of 20\% in classification accuracy and 9.7 in diversity. This confirms our choice to use this configuration for Interformer. Table ~\ref{ablationmultiheadK3HI} shows the same ablation for the K3HI dataset and we observe the same behavior as for SBU except for the diversity  where other configurations have lower values than using both multihead attention. We believe this to be due to the more noisy nature of the K3HI dataset, which deteriorates the diversity measures.

\begin{table}[!t]
\centering
\small
\begin{tabular}{@{}c c| c c @{} } 
\toprule
T multihead & S multihead & Accuracy $\uparrow$ & Diversity $\downarrow$ \\
\midrule
 -&-  & 60.0 &10.6\\ 
\checkmark &-  & 70.0 &2.0\\ 
 -& \checkmark & 60.0 &5.1\\ 
 \checkmark&\checkmark & \textbf{80.0} & \textbf{0.9} \\

 \bottomrule
 \end{tabular}
\caption{Ablation study of  Interformer on the SBU dataset.}
\label{ablationmultihead} 
\end{table}

\begin{table}[!t]
\centering
\small
\begin{tabular}{@{}c c| c c @{} } 
\toprule
T multihead & S multihead & Accuracy $\uparrow$ & Diversity $\downarrow$ \\
\midrule
 -&-  & 21.4 &11.0\\ 
\checkmark &-  & 42.6 &\textbf{5.3}\\ 
 -& \checkmark & 35.7 &5.7\\ 
 \checkmark&\checkmark & \textbf{46.4} & 13.7 \\

 \bottomrule
 \end{tabular}
\caption{Ablation study of  Interformer on the K3HI dataset}.
\label{ablationmultiheadK3HI} 
\end{table}

\section{Limitations}
InterFormer presents two  main limitations: (i) Due to the huge variability of complex motions, it is hard to stay true to the ground truth, making it difficult to evaluate the results in these cases; 
(ii) We are able to generate realistic motion for long sequences (tested up to 40 seconds)
To do this we cut the action sequence into smaller sub-sequences that we use for generation. We then generate all these sequences the same way as we do for shorter sequences. Only for the second sub-sequence onward the first frame used to give the initial position does not come from ground truth but instead is the last generated frame from the previous sub-sequence. We can see in ``DuetDance-long.mp4'' from our supplementary material that this way InterFormer is able to generate reaction sequences for long motion. However, due to the accumulation of errors over time, the generation diverges more and more from the ground truth up to the point where it is hard to know how much action is taken into account in the generation. It is even more true that very long motions are usually complex ones, which means we also face the first limitation.

\section{Conclusion}
We present InterFormer, a novel human reaction generation Transformer. InterFormer is the first Transformer architecture used to solve the problem of human reaction generation challenge. InterFormer consists of four modules:  a motion encoder, a motion decoder, a skeleton adjacency module, and an interaction distance module. The ablation study on SBU has shown the effectiveness of the four components of the InterFormer. We have both qualitatively and quantitatively evaluated our reaction generation framework. The results show that InterFormer outperforms state-of-the-art approaches in terms of FVD, classification, and diversity score on three challenging datasets SBU, K3HI, and DuetDance. The qualitative results show also the ability of InterFormer to generate realistic human reactions. Interformer is a deterministic approach. Although we have proposed an approach to mitigate this problem, the diversity of responses generated remains limited and should be improved. It is still difficult to generate  complex human motion. Although our results on  the dance dataset show that we are able to generate dance movements, we are still not able to generate more subtle motions present in the dataset. The lack of large interaction datasets makes it difficult to evaluate feedback generation. Although large interaction datasets exist, such as some classes of NTUs, they are not annotated to separate action from reaction motion.
It is difficult to evaluate the performance on long-term motion due to the lack of 
appropriate data.

\section*{Acknowledgments}
This project has received financial support from the CNRS through the 80—Prime program, from the French State, managed by the National Agency for Research (ANR) under the Investments for the future program with reference ANR- 16-IDEX-0004 ULNE and by the EU H2020 project AI4Media
under Grant 951911. Most of this work was done when Naima Otberdout was at University of Lille.






\begin{IEEEbiography}[{\includegraphics[width=1in,height=1.25in,clip,keepaspectratio]{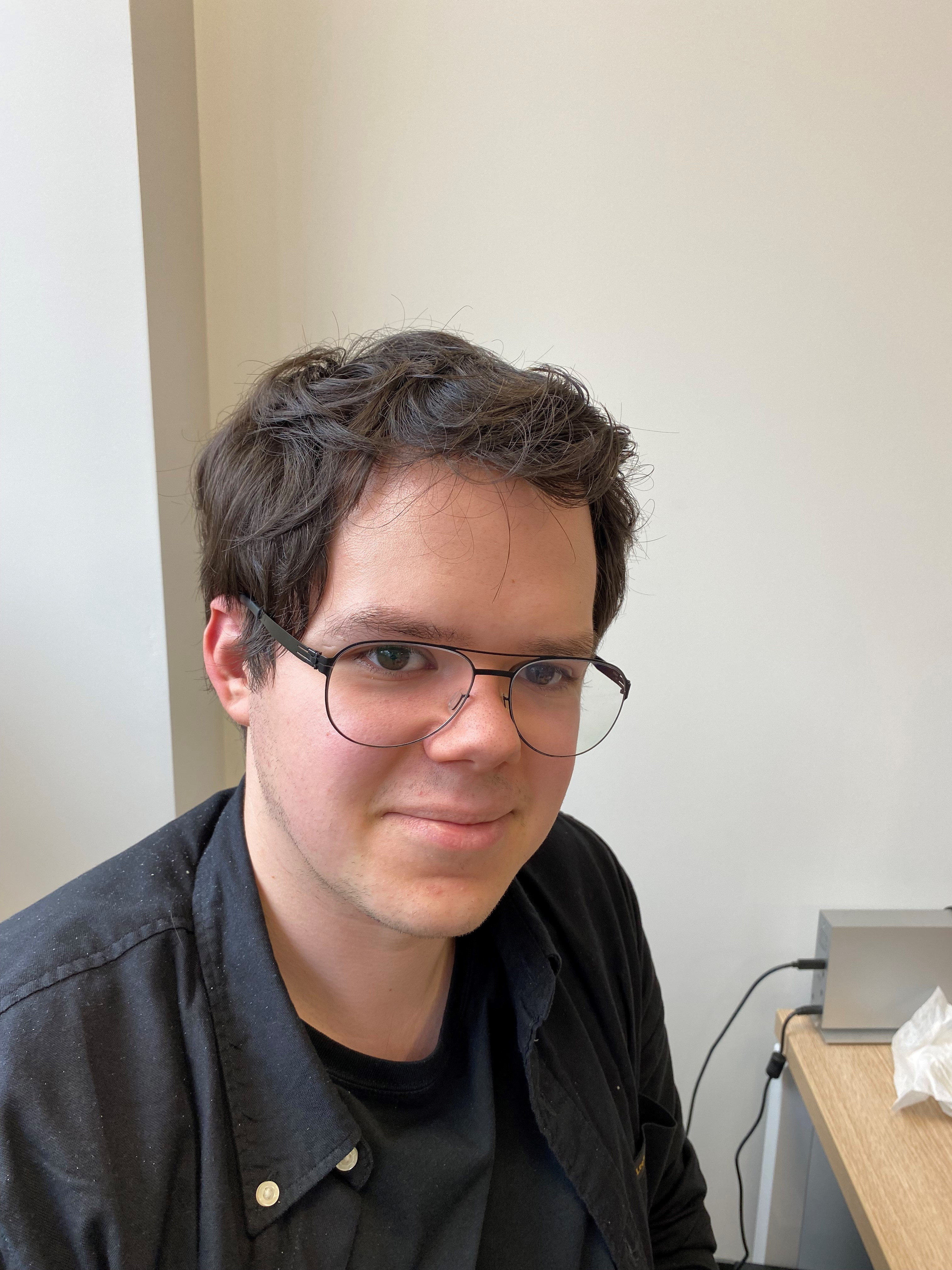}}]
{Baptiste Chopin} received the Engineering degree in computer science from IMT Nord Europe (France).  He is currently pursuing the Ph.D. degree with the university of Lille (France). His research concern computer vision and the generation of human motion with application to cognitive sciences.
\end{IEEEbiography}

\begin{IEEEbiography}[{\includegraphics[width=1in,height=1.25in,clip,keepaspectratio]{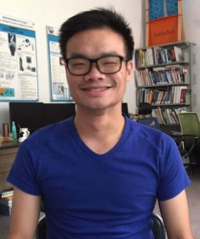}}]{Hao Tang} is currently a Postdoctoral with Computer Vision Lab, ETH Zurich, Switzerland.
He received the master’s degree from the School of Electronics and Computer Engineering, Peking University, China and the Ph.D. degree from the Multimedia and Human Understanding Group, University of Trento, Italy.
He was a visiting scholar in the Department of Engineering Science at the University of Oxford. His research interests are deep learning, machine learning, and their applications to computer vision.
\end{IEEEbiography}

\begin{IEEEbiography}[{\includegraphics[width=1in,height=1.25in,clip,keepaspectratio]{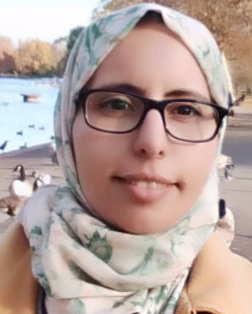}}]
{Naima Otberdout} received the master’s degree in computer sciences and telecommunication from Mohammed V University, Rabat, Morocco in 2016. She received the Ph.D. degree in computer science from the same university in 2021. After a post-doctoral position in the University of Lille in France, she is currently a Research and Education Fellow in Ai movement - Mohammed VI Polytechnic University in Morocco. 
\\ Her current research interests include computer vision and pattern recognition with applications to human behavior understanding.
\end{IEEEbiography}

\begin{IEEEbiography}[{\includegraphics[width=1in,height=1.25in,clip,keepaspectratio]{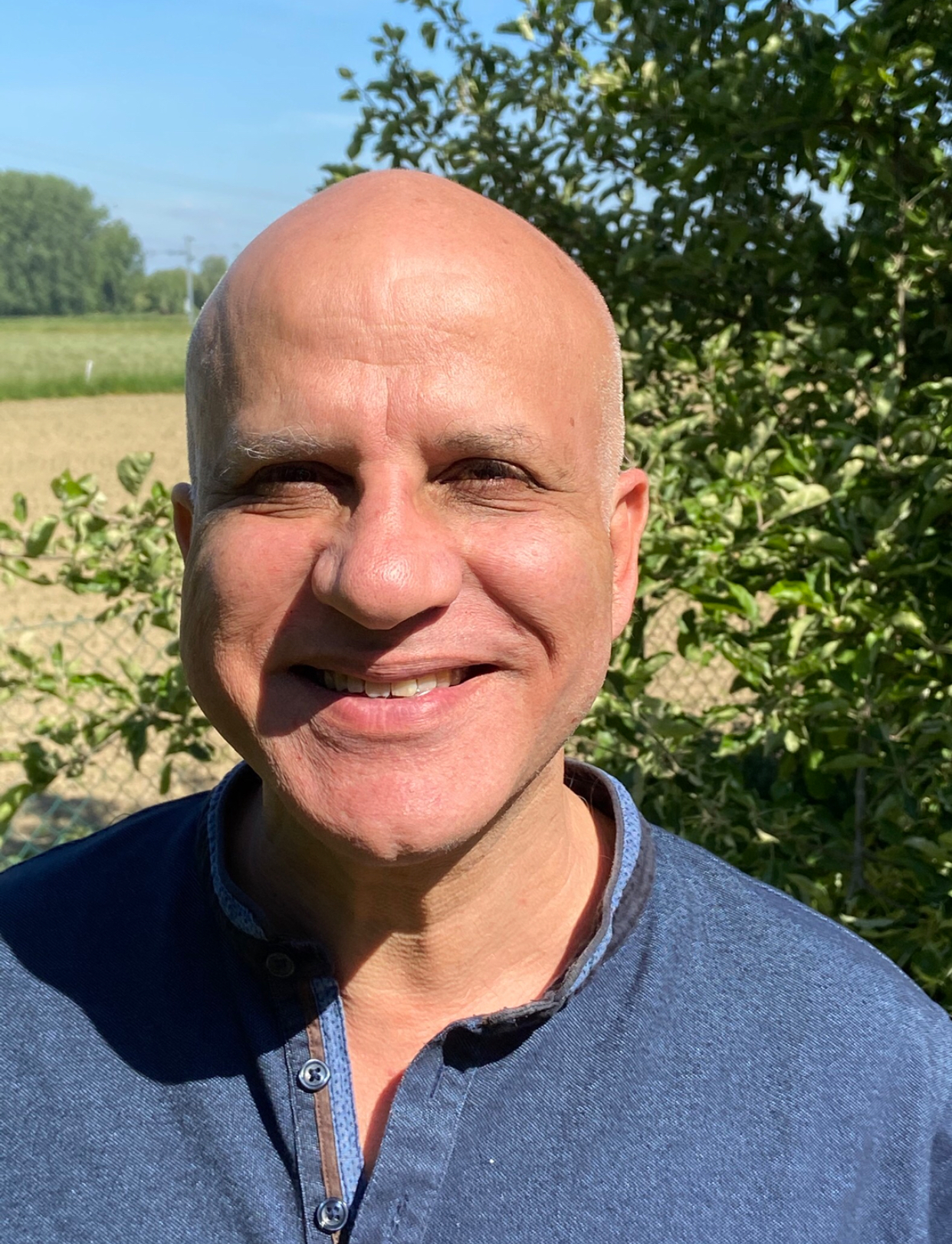}}] {Mohamed Daoudi} is Full Professor of Computer Science at IMT Nord Europe and the lead of Image group at CRIStAL Laboratory (UMR CNRS 9189). His research interests include pattern recognition, shape analysis and computer vision. He has published over 150 papers in some of the most distinguished scientific journals and international conferences. He is Associate Editor of Image and Vision Computing Journal, IEEE Trans. on Multimedia, Computer Vision and Image Understanding, IEEE Trans. on Affective Computing and Journal of Imaging. He has served as General Chair of IEEE International Conference on Automatic Face and Gesture Recognition, 2019. He is a fellow of the International Association for Pattern Recognition.
\end{IEEEbiography}

\begin{IEEEbiography}[{\includegraphics[width=1in,height=1.25in,keepaspectratio]{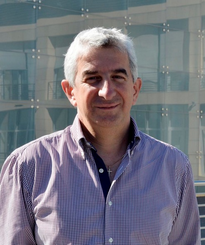}}]{Nicu Sebe} is Professor with the University of Trento, Italy, leading the research in the areas of multimedia information retrieval and human behavior understanding. He was the General Co-Chair of ACM Multimedia 2013, and the Program Chair of ACM Multimedia 2007 and 2011, ECCV 2016, ICCV 2017 and ICPR 2020. He is a fellow of the International Association for Pattern
Recognition.
\end{IEEEbiography}

\vfill

\end{document}